\documentclass[letterpaper]{article} 
\usepackage{aaai25}  
\usepackage{times}  
\usepackage{helvet}  
\usepackage{courier}  
\usepackage[hyphens]{url}  
\usepackage{graphicx} 
\usepackage{natbib}  
\usepackage{caption} 
\usepackage{algorithm}
\usepackage{algorithmic}
\usepackage{amsmath}
\usepackage{amsfonts}
\usepackage{multirow}
\usepackage{booktabs}

\usepackage{newfloat}
\usepackage{listings}
\DeclareCaptionStyle{ruled}{labelfont=normalfont,labelsep=colon,strut=off} 
\floatstyle{ruled}
\newfloat{listing}{tb}{lst}{}
\floatname{listing}{Listing}
\nocopyright 

\title{Text-Video Multi-Grained Integration for Video Moment Montage}
\def\task{Video Moment Montage}
\author {
    Zhihui Yin\textsuperscript{\rm 1}\thanks{Work done during an internship at Kuaishou Technology},
    Ye Ma\textsuperscript{\rm 2},
    Xipeng Cao\textsuperscript{\rm 2},
    Bo Wang\textsuperscript{\rm 2},
    Quan Chen\textsuperscript{\rm 2}\thanks{Corresponding author.},
    Peng Jiang\textsuperscript{\rm 2}
}
\affiliations {
    \textsuperscript{\rm 1} Xidian University\\
    \textsuperscript{\rm 2} Kuaishou Technology\\
}

\usepackage{bibentry}

\begin{document}

\maketitle

\begin{abstract}

The proliferation of online short video platforms has driven a surge in user demand for short video editing. However, manually selecting, cropping, and assembling raw footage into a coherent, high-quality video remains laborious and time-consuming. To accelerate this process, we focus on a user-friendly new task called Video Moment Montage (VMM), which aims to accurately locate the corresponding video segments based on a pre-provided narration text and then arrange these video clips to create a complete video that aligns with the corresponding descriptions. The challenge lies in extracting precise temporal segments while ensuring intra-sentence and inter-sentence context consistency, as a single script sentence may require trimming and assembling multiple video clips. To address this problem, we present a novel \textit{Text-Video Multi-Grained Integration} method (TV-MGI) that efficiently fuses text features from the script with both shot-level and frame-level video features, which enables the global and fine-grained alignment between the video content and the corresponding textual descriptions in the script. To facilitate further research in this area, we introduce the Multiple Sentences with Shots Dataset (MSSD), a large-scale dataset designed explicitly for the VMM task. We conduct extensive experiments on the MSSD dataset to demonstrate the effectiveness of our framework compared to baseline methods. 

\end{abstract}

\section{Introduction}
\label{sec:intro}

Short videos have emerged as the dominant medium for sharing information, daily life moments, and advertisements on online platforms, spurring a growing demand among platform users for short video editing and montage creation. High-quality short videos typically require the skillful arrangement of carefully selected and trimmed footage segments, often accompanied by a narration script. However, learning video editing tools and curating media assets for video compilation is time-consuming and labor-intensive for the general public. In contrast, the intelligent generation of short video montages based on self-created narration scripts and a library of video assets can significantly reduce the cost and complexity of video production.
\begin{figure}[!ht]
\centering
\includegraphics[width=0.47\textwidth]{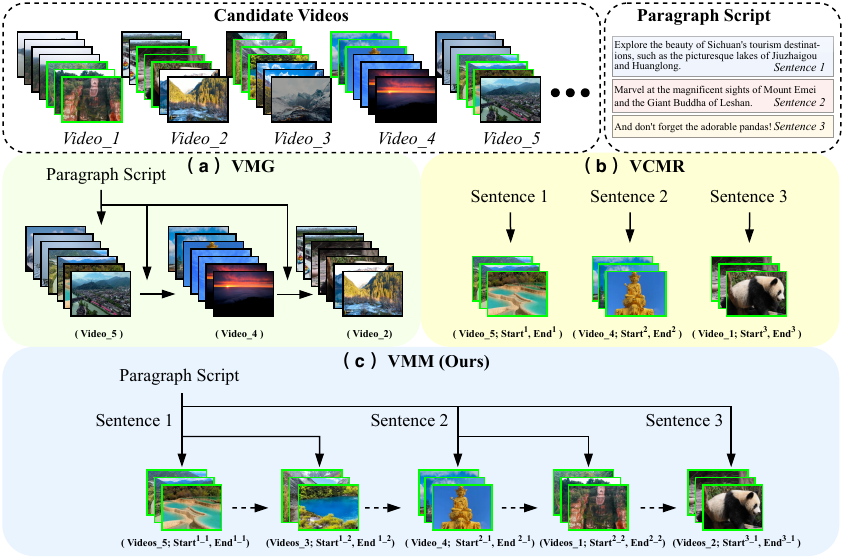}
\caption{Differences between VMM and other tasks. (a) VMG involves creating video montages using complete videos. (b) VCMR utilizes individual sentences to retrieve specific moments within a video corpus but cannot form a coherent script-level video. (c) The key distinction of our proposed VMM task is that the video montage is composed of selected moments from the videos, and a single sentence can correspond to the composition of multiple video fragments, enabling more flexible video construction.}
\label{fig:task}
\end{figure}

In recent years, there have been primary attempts to tackle text-to-video retrieval-based video montage generation (VMG)~\cite{truong2016quickcut,wang2019write,xiong2022transcript,yang2023shot}. These methods retrieve relevant videos or shots from a large video gallery based on user-specified keywords~\cite{wang2019write} or query sentences~\cite{xiong2022transcript,yang2023shot}, and then sequentially assemble them to create the final video montage. However, while these retrieval-based methods have provided a foundational step forward, the videos produced are weakly aligned with the scripts, neglecting the fine-grained correspondence between sentences and segments in videos. Moreover, these methods treat video candidates as a fixed set without accurately localizing video shots to specific moment segments. 


The Video Moment Retrieval (VMR)~\cite{gao2017tall,anne2017localizing,lei2021qvhighlights,moon2023query} and Video Corpus Moment Retrieval (VCMR)~\cite{zhang2021video,query_aware_ranking,selective_query} techniques have established connections between textual descriptions and corresponding video moments, using sentence-level queries to retrieve relevant temporal segments from single or multiple videos. It is worth noting that these methods focus on single-sentence video retrieval, and the textual information utilized in these approaches tends to be more oriented towards content descriptions rather than the narration scripts that are more useful for short-form video montage applications.

To this end, we propose a novel task that better aligns with real-world short video editing requirements, termed Video Moment Montage (VMM), as illustrated in Figure~\ref{fig:task}. It involves a narration script with multiple sentences and a set of candidate videos the users provide. The task aims to automatically locate suitable video segments from the source videos for each script sentence and merge them. VMM presents two key challenges: 1) retrieving the corresponding moments in the videos based on the given text scripts, and 2) enabling the model to learn the fine-grained contextual relationships between individual sentences in the script and their corresponding video clip segments and the global-grained contextual relationships between collections of sentences and their associated sets of video clip segments.



To address this task, we propose \textit{Text-Video Multi-Grained Integration} (TV-MGI) that facilitates the multi-level fusion of text features and video features for fine-grained multimodal understanding and prediction. Specifically, we introduce a frame-shot-text feature fusion mechanism to capture the relationship between different modalities at different levels. These fused features are then used to predict the corresponding video segments for each sentence. This design enables our model to handle videos from diverse sources, achieving accurate cross-modal alignments for precise predictions. Furthermore, we propose a synthetic data augmentation module that simulates realistic video materials containing redundant information, which results in models trained with enhanced generalization capabilities. 

Existing video datasets~\cite{lei2021qvhighlights,yang2023shot} lack multi-sentence scripts or fine-grained annotations aligning sentences with video segments, rendering them unsuitable for the VMM task. Hence, we collect a large-scale, fine-grained aligned text-video dataset called Multiple Sentences with Shots Dataset (MSSD), with extracted multi-sentence scripts and corresponding video segment annotations. Besides, we carefully construct the test set for MSSD, where each instance comprises a script, the candidate videos, and the well-edited final short video montage.

In summary, the main contributions of this work include:
\begin{itemize}
    \item We introduce \task, a novel task that seeks to automate the selection of video segments and the cutting from raw videos based on narration scripts, streamlining the creation of video montage.
    \item We propose a Text-Video Multi-Grained Integration method that adopts a frame-shot-text feature fusion mechanism for detailed multimodal understanding and high-quality montage generation. 
    \item We collect a large-scale dataset called MSSD for \task. Experiments on the dataset demonstrate the superiority of our proposed framework compared to reproduced baselines.
\end{itemize}

\section{Related Work}

\subsection{Video Montage Generation}


The concept of creating video montages based on textual scripts has a long history~\cite{chua1995video,ahanger1998automatic}. Recent advancements and interest in this area have been rejuvenated~\cite{truong2016quickcut,leake2017computational,wang2019write,xiong2022transcript,yang2023shot} due to the growth of social media. These methods retrieve shots from a gallery based on text inputs and assemble them accordingly. \cite{wang2019write} match keywords, \cite{xiong2022transcript} retrieve shots for query sentences, and \cite{yang2023shot} extend this to multiple sentences. In contrast, our task requires each script sentence to be synchronized in time and content with its video segments, aligning with short video editing needs.

\subsection{Video Moment Retrieval}

The task of VMR localizes relevant moments in a video using a natural language query. Existing works focus on scenarios where a single video moment corresponds to a given query, using either proposal-based~\cite{anne2017localizing,gao2017tall,zhang20202dtan,li2024momentdiff} or proposal-free~\cite{zhang2020vslnet,cao-etal-2021-pursuit,liu2022umt} methods. Proposal-based approaches typically generate candidate segments and select the best-matching one, whereas proposal-free approaches directly predict timestamps without the need for candidate generation. In addition,~\cite{bao2021dense,jiang2022semi,tan2023hierarchical} explore the setting where multiple consecutive video segments are localized according to a paragraph query. However, our VMM task goes beyond sentence-level moment retrieval, requiring the arrangement and composition of multiple fine-grained short video moments corresponding to each sentence in the reference text.




\begin{figure*}[!t]
\centering
\includegraphics[width=\textwidth,trim=0 60 0 0,clip]{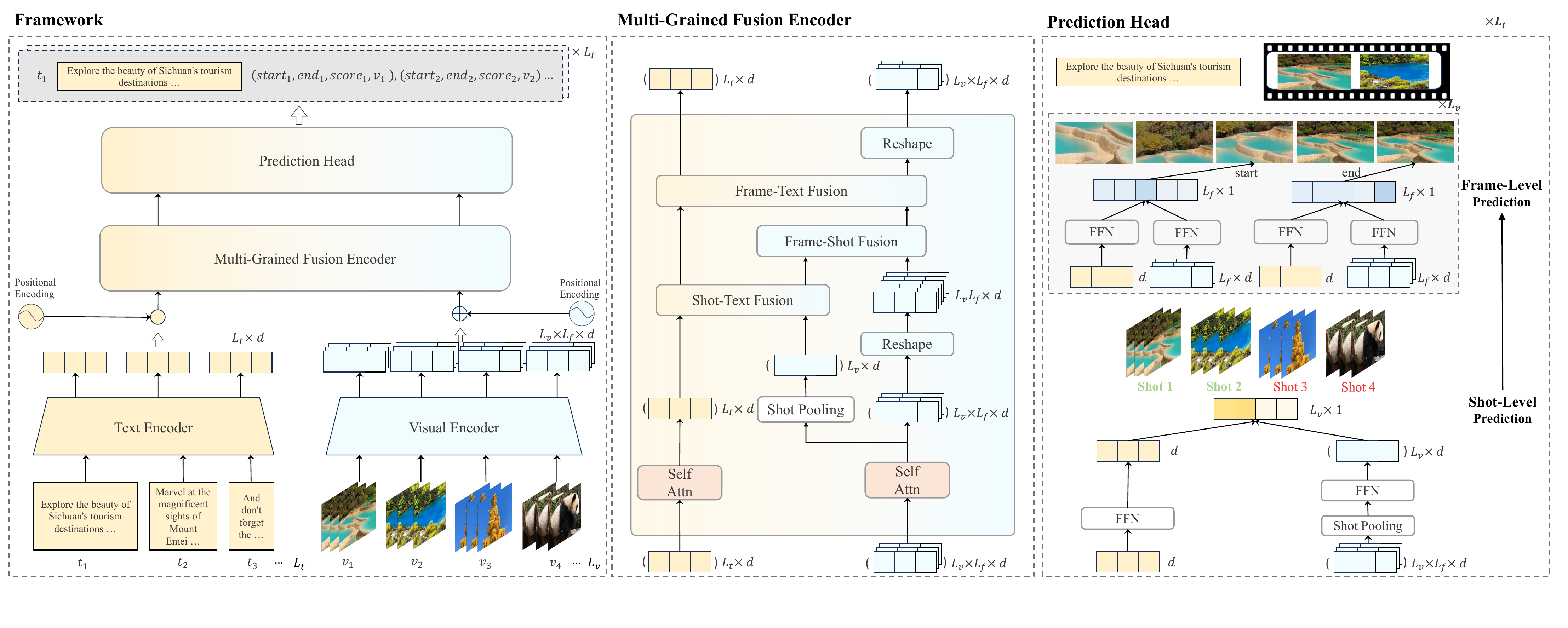} 
\caption{The illustration of basic architectures of our proposed TV-MGI. From left to right, there are the overall framework, multi-grained fusion module, and the prediction head. We first employ visual and text encoders to map the input text and video into embeddings. Next, we apply attention fusion to the visual and textual features at both the shot- and frame-level. The output representations from the fusion encoder are utilized for prediction at both the shot- and frame-level, enabling the generation of multiple video segments corresponding to each sentence in the script.}
\label{fig:model}
\end{figure*}

\subsection{Video Corpus Moment Retrieval}
Video Corpus Moment Retrieval (VCMR) aims to identify relevant video segments from a large corpus for a given textual query. This task was initially introduced by~\cite{victor2019temporal} and involves two sub-tasks: video retrieval and temporal localization. Existing VCMR approaches can be single-stage,~\cite{lei2020tvr,li2020hero,zhang2021video,yoon2022selective} using a joint model, or two-stage~\cite{query_aware_ranking,zhang2023video,chen2023cross,hou2024improving}, with dedicated modules for each sub-task. For instance, single-stage methods like~\cite{lei2020tvr} combine video retrieval and localization, while two-stage methods~\cite{zhang2023video, chen2023cross, hou2024improving} utilize a pre-trained retrieval head and introduce additional components for localization. Our works focus on the VMM task, which not only necessitates the extraction of cross-modal relevance but also requires the consideration of contextual relationships between and within sentences in the reference script without the aid of video subtitles.



\section{Method}

The VMM task matches and localizes video segments for all sentences in a multi-sentence script. It aligns the textual description with the visual content, creating a coherent and contextually relevant short video montage. The script is represented as a list $\mathcal{T} = \{t_1, t_2, \ldots, t_{L_t}\}$ where $t_i$ denotes the $i$-th sentence in the script and $L_t$ represents the total number of sentences. The input shots are represented as $\mathcal{V} = \{v_1, v_2, \ldots, v_{L_v}\}$ where $v_j$ refers to the $j$-th shot and $L_v$ is the total number of shots. Each $v_j$ contains all the extracted frames for the shot, which can be formalized as $v_j = \{f^j_1, f^j_2, \ldots, f^j_{L_f}\}$, where ${L_f}$ represent the number of frames within the shot. The shot boundary extracted by Transnet v2~\cite{souvcek2020transnet} denoted as $\mathcal{C} = \{(s_1, e_1), (s_2, e_2), \ldots, (s_{L_v}, e_{L_v})\}$, where $ 1 \leq s_k < e_k \leq L_vL_f $. They are used to identify all shots and their order. In this way, we can establish a time alignment between the sentences and shots by comparing the timestamps of the sentences and shots, which indicates the matching relationship between them as $\mathcal{M} = \{ (t_i, U_i) | t_i \in \mathcal{T}, U_i \subseteq U  \}$, where $U = \{(start, end, rank, l_v) | 1 < start \leq end \leq L_f \& 1 < l_v \leq L_v \} $ represents the set of segments where each shot spans from the $start$ frame to the $end$ frame. The $rank$ denotes the segment's position within the sentence $t_i$, ranging from 1 to the size of $U_i$. A higher rank indicates an earlier appearance of the shot in time, while $l_v$ represents the corresponding shot for that segment.
It is noteworthy that $s_k$ and $e_k$ denote the absolute frame positions, while $start$ and $end$ refer to the relative positions within a shot.

The overall architecture of the model is demonstrated in Figure~\ref{fig:model}. We first employ a text encoder and a visual encoder for the given shots and script inputs to extract their features. These extracted features are then fed into a multi-grained fusion encoder. Within each layer of the fusion encoder, we first aggregate the frame-level video features to form shot-level video features, followed by multiple rounds of fusion between the different video and text features. Finally, a prediction head is applied to the outputs of the fusion encoder to carry out multi-granularity predictions at both the frame-level and shot-level, constructing the ultimate results.
\subsection{Feature Extraction}

For the input shots $\mathcal{V} = \{v_1, v_2, \ldots, v_{L_v}\}$ consisting of uniformly sampled frames, we utilize a pre-trained CLIP~\cite{radford2021clip} visual encoder to encode each frame to extract frame-level visual features $Y = \{y^j_1,y^j_2,\ldots,y^j_{L_f}\}_{j=1}^{L_v} \in \mathbb{R}^{L_v \times L_f\times{d}}$. 
For the input script $\mathcal{T} = \{t_1, t_2, \ldots, t_{L_t}\}$, we use the text encoder from CLIP to extract sentence-level text features $X = \{x_i\}_{i=1}^{L_t} \in \mathbb{R}^{L_t\times{d}}$. Note that all the video and text features are mapped to the same dimensionality $d$ through a linear projection. We employ CLIP for frame-level and sentence-level feature extraction due to its robust and well-aligned cross-modal features.

\subsection{Multi-grained Fusion Encoder}

We propose a multi-grained fusion encoder to integrate features across different semantic levels. The fusion encoder consists of $N$ encoder layers. We use self-attention blocks in each layer to capture intra-modal relationships separately for frame-level and text features. Since attention is permutation-invariant, we add fixed positional encodings to the input features in each layer to preserve the order of sentences and video frames. The resulting features are still denoted as $X$ and $Y$ for simplicity. Then we compute the shot-level features $Z = \{z_j\}_{j=1}^{L_v}$ as the mean pooling of the frame-level features within each shot:
\begin{equation}
    z_j = \mathrm{MeanPooling}(\{y^j_f\}_{f=1}^{L_f})
\end{equation}

We then proceed with three sequential types of feature fusion: shot-text, frame-shot, and frame-text fusion. 
We use cross-modality multi-head attention~(X-MHA)~\cite{li2022glip} as the fusion module. In X-MHA, each head computes the context representations of one modality by attending to anothor modality. We take frame-text fusion as an example. Specifically, given the text features $X$ and the reshaped frame-level visual features $Y' \in \mathbb{R}^{L_vL_f\times{d}}$, X-MHA computes their contextual representations $[O_X, O_Y'] = \mathrm{X}\text{-}\mathrm{MHA}(X, Y')$ with two simultaneous cross-attention operations as:
\begin{equation}
\begin{aligned}
A &= {XW_X^q \cdot ({{Y'}W_{Y'}^q})^{\top}}/{\sqrt{d}} \\
O_X &= \mathrm{softmax}(A)XW_X^{v}W_X^{o} \\
O_{Y'} &= \mathrm{softmax}(A^{\top})YW_{Y'}^{v}W_{Y'}^{o} \\
\end{aligned}
\end{equation}
where $W_{*}^q$, $W_{*}^v$ and $W_{*}^o$ represent the trainable weights for the query, value, and output linear transformations, respectively. The processes for frame-shot fusion and shot-text fusion are similar. The final frame-level features $Y^{ft}$ and text features $X^{ft}$ are obtained through the following multi-grained video-text fusion process:
\begin{equation}
\begin{aligned}
\mathrm{[}Z^{vt}, X^{vt}\mathrm{]} &= \mathrm{X}\text{-}\mathrm{MHA}\left(Z, X\right),\; \text{Shot-Text}\\
[Z^{fv}, Y^{fv}] &= \mathrm{X}\text{-}\mathrm{MHA}\left(Z^{vt}, Y'\right),\; \text{Frame-Shot} \\
[Y^{ft}, X^{ft}] &= \mathrm{X}\text{-}\mathrm{MHA}\left(Y^{fv}, X^{vt}\right),\; \text{Frame-Text} \\
\end{aligned}
\end{equation}

Note that such a process happens in each layer of the fusion encoder, which brings a deep fusion between features at different semantic levels. The output of this module is the fused visual representation $\tilde{Y}\in \mathbb{R}^{L_v\times L_f\times{d}}$ and textual representation $\tilde{X}\in \mathbb{R}^{L_t\times {d}}$ after feature fusion.

\subsection{Prediction and Loss}
In short video editing, editors typically select and order shots based on the script and available footage and choose specific segments from those shots.
Following this, the prediction step involves two steps: identifying and ranking shots corresponding to a given sentence in the script and localizing fine-grained video segments within each related shot.

Formally, given the output frame-level features $\tilde{Y} $ and text features $\tilde{X}$, we first aggregate the shot-level features $\tilde{Z}$ using the same pooling operation that was previously introduced. For the shot-level predictions, we compute a shot-text matching scores $p_{ik}^{m}$ for each shot-level feature $z_k \in \tilde{Z}$ and sentence feature $x_i \in \tilde{X}$ as:
\begin{equation}
    p_{ik}^{m} = \sigma\left(\left(W_m^{t} \cdot x_i\right) \cdot \left(W_m^{v} \cdot z_k\right)^{\top}\right)
\end{equation}
where $W_m^{t}$ and $W_m^{c}$ are trainable projection weights and $\sigma$ represents the sigmoid function. During training, we apply the binary cross-entropy loss as:
\begin{equation}
    \mathcal{L}_{m} = -\sum\limits_{i=1}^{L_t}\sum\limits_{k=1}^{L_v} m_{ik}\log(p_{ik}^{m}) + (1-m_{ik})\log(1-p_{ik}^{m})
\end{equation}
where $m_{ik}$ is set to $1$ if the corresponding sentence and shot are matched, and $0$ otherwise.

To capture the ordinal relationship between shots, we draw inspiration from the rank-aware contrastive loss~\cite{hoffmann2022ranking}, which facilitates contrasting learning of the ranking order.
In our task, we aim for higher predicted scores $p_{ik}^{m}$ for shots with higher $rank$ score in text $t_i$. For each matching relationship $(t_i, U_i)$ where $L_u$ denotes the size of $U_i$, we iterate all shots for $L_u$ times. In each iteration, shots with higher $rank$ scores than the iteration index ($r \in {0,1,\ldots,L_u-1}$) as the positive examples $Y^{pos}_{r}$, while the remaining shots are utilized as negative examples as the negative set $Y^{neg}_{r}$. The rank-aware contrastive loss $L_r$ is defined as:
\begin{equation}
    \mathcal{L}_{r} = -\sum\limits_{i=1}^{L_t}\sum\limits_{r=0}^{L_u-1}log\frac{\sum_{y\in Y^{pos}_{r}}exp(p_{iy}^m/\tau)}{\sum_{y\in Y^{pos}_{r} \bigcup Y^{neg}_{r}}exp(p_{iy}^m/\tau)}
\end{equation}
where $\tau$ is a temperature scaling parameter. It should be note that $Y^{neg}_{r}$ also include all shots which are not in $U_i$.

For the frame-level predictions, we estimate the probabilities of each frame within a matched shot as either the starting or ending frame of the final video segment. Similar to the shot-text matching score, we get the start and end scores for $f^j_{k} \in v_j$ as $p_{ijk}^{s}$ and $p_{ijk}^{e}$, respectively. Instead of using the sigmoid function, these scores are normalized within the shot using the softmax function. We compute the start and end loss as:
\begin{equation}
    \mathcal{L}_{g} = -\sum\limits_{i=1}^{L_t}\sum\limits_{j=1}^{L_v}\sum\limits_{k=1}^{L_f} \log(p_{ijk}^{s}) + \log(p_{ijk}^{e})
\end{equation}

The overall objective of our TV-MGI for VMM is the sum of the shot-level and frame-level losses:
\begin{equation}
    \mathcal{L} = \mathcal{L}_{m} + \mathcal{L}_{r} + \mathcal{L}_{g}
\end{equation}

\subsection{Synthetic Data Augmentation}
Since training directly on the original video's script and shots unavoidably introduces bias in the inference phase, we mitigate this issue by employing synthetic data augmentation to enhance the model's generalization capability. 

We first introduce the \textbf{Shuffling} strategy. This involves randomly shuffling all shots in the original video $\mathcal{V}$ to create synthetic video materials. By doing so, we mimic the unordered nature of raw video materials, where video segments can appear at random positions throughout the original footage. Secondly, we utilize the \textbf{Injection} strategy. This strategy involves randomly inserting shots from other videos within the same cluster into $\mathcal{V}$. The purpose of this method is twofold: to simulate the presence of redundant parts commonly found in raw materials and to introduce challenging negative examples that enrich the training data. We compare the performance of different data construction strategies in the ablation study. For more details on data construction, please refer to our supplementary materials.

We only need to input multiple raw materials sets and the textual script during the actual usage phase. The model can automatically match each sentence with the corresponding materials and arrange them correctly based on the script's context. Finally, the model can locate specific segments of the matched materials, achieving end-to-end alignment between the raw materials and the script.
\begin{figure}[!t]
        \begin{minipage}[t]{0.2\linewidth}
		\centering
		\includegraphics[width=0.7in]{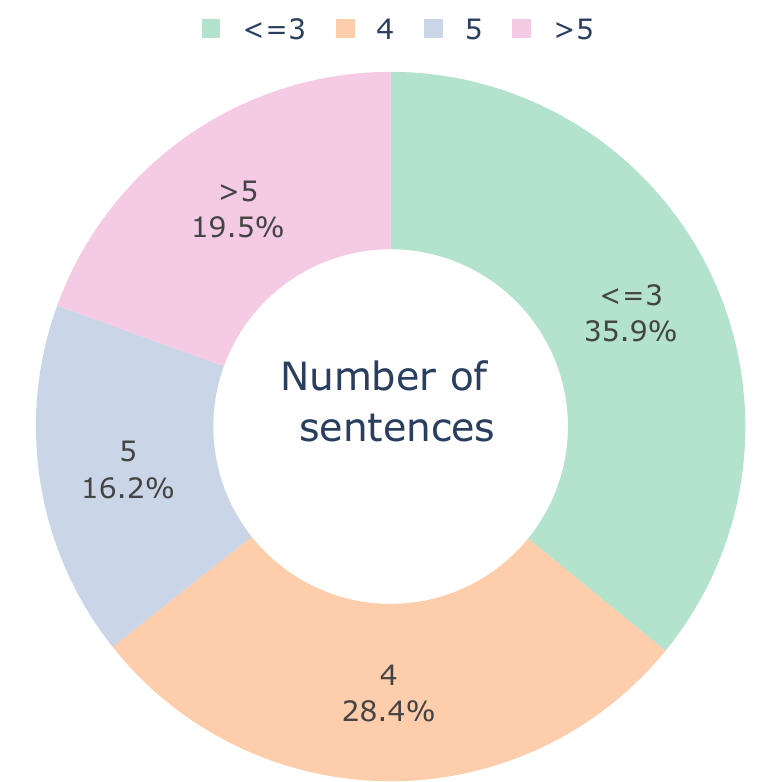}
	\end{minipage}
         \begin{minipage}[t]{0.2\linewidth}
            \centering
            \includegraphics[width=0.7in]{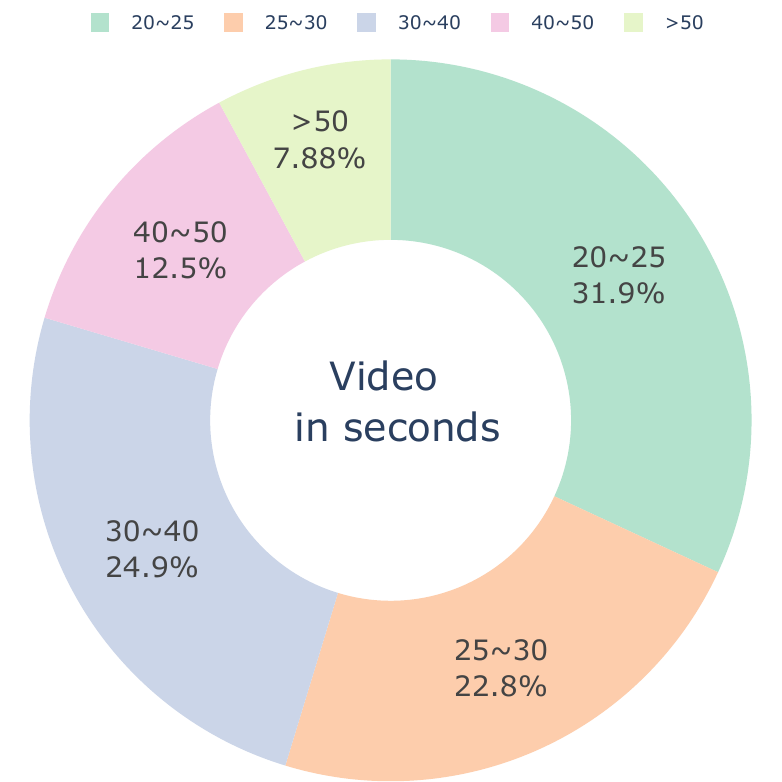}
        \end{minipage}
        \begin{minipage}[t]{0.5\linewidth}
        		\centering
        		\includegraphics[width=1.9in]{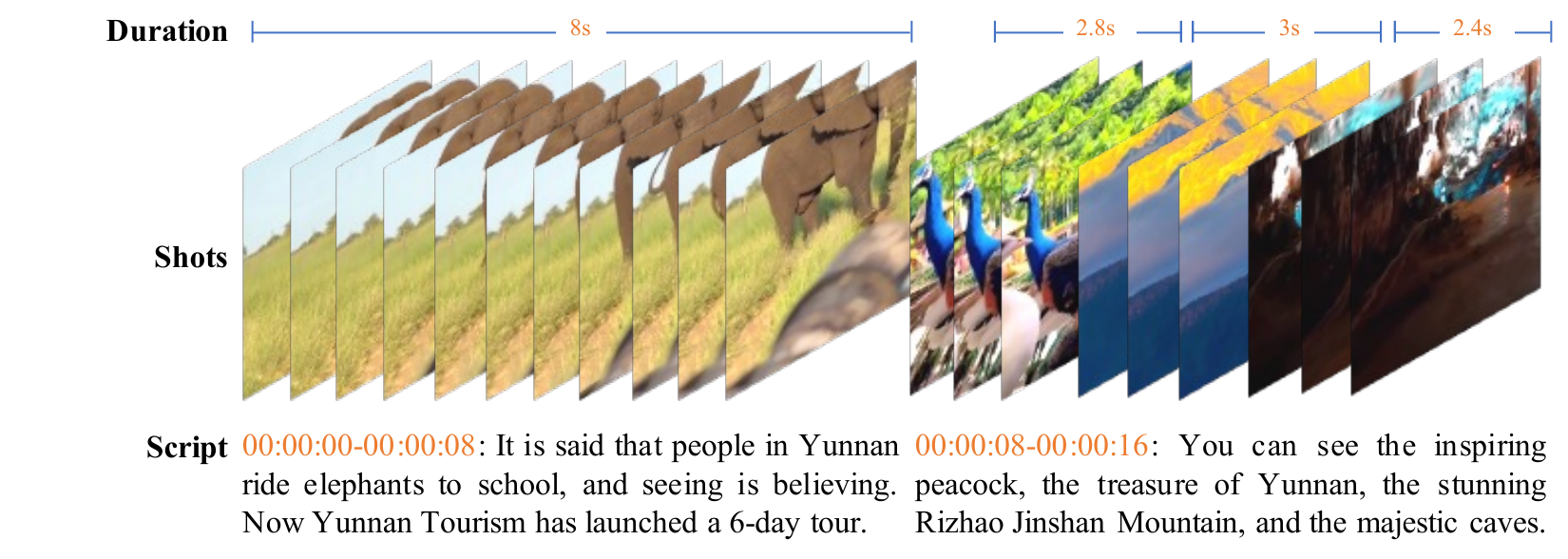}
	\end{minipage}
	\caption{\textbf{Left}. The statistics of scripts and videos contained in our training dataset. \textbf{Right}. Displaying a sample of dataset, including the script, shots, and the temporal alignment between visuals and script timeline. More dataset details and cases are in the supplementary materials.}
	\label{fig:statistic}
\end{figure}

\section{Dataset Construction}

A dataset for the VMM task should consist of multiple sentences in a script, temporally aligned with multiple candidate segments. The previously available datasets for video montage generation, video moment retrieval, or video corpus moment retrieval do not meet the requirement. Consequently, we construct a new large-scale script-video dataset (Multiple Sentences with Shots Dataset, MSSD) collected from a Chinese short video platform. We initially utilize automatic speech recognition (ASR) to extract video narration scripts and timestamps at the sentence and word levels. However, ASR results can be noisy due to background music and recognition errors. We further utilize optical character recognition (OCR) to enhance accuracy in recognizing video subtitles. This helps us refine the ASR results and retain videos that have manually created subtitles. We translate the Chinese text to make the data more universally applicable and provide the translated English version in all figures. The MSSD dataset consists of 187,351 videos and narration scripts, totalling 1,625 hours of video and 808,936 sentences. We leverage Transnet v2~\cite{souvcek2020transnet} to extract the start and end frame information for all shots in the videos. We carefully align the sentences with the corresponding shots in the timeline. This process ensures that the sentences' timing matches the video's visual content. We show more statistics of this training dataset and a sample from this training dataset in Figure~\ref{fig:statistic}. 

For evaluation, we create a test set that simulates real short video montage scenarios with a script comprising multiple materials. We first employ clustering techniques on all videos, leveraging their category tags and visual content representation. We randomly sample 2,921 clusters, each containing two randomly selected videos. Subsequently, we utilize the shots from these two videos to perform synthetic data augmentation as candidate materials. One of the original videos will be selected to serve as the ground truth. The model's primary objective is to identify suitable segments from the candidate shots that align with the provided script, aiming to maximize the similarity between the selected segments and the ground truth video. This benchmark enables a comparative evaluation between the model's output and the short video montage results generated by human editors, providing valuable insights into the model's performance. Please refer to the supplementary materials for a more comprehensive dataset description, including its composition and characteristics.

\section{Experiments}

\begin{table*}[t!]
    \setlength{\tabcolsep}{0.4em}
    \small
    \centering
    \captionsetup{skip=1pt}
    \resizebox{\textwidth}{!}{\begin{tabular}{lcccccccccccccccccc}
    \toprule
    \multirow{2}{*}{ Method } & \multicolumn{2}{c}{R1} & \multicolumn{3}{c}{mAP@5} & \multicolumn{2}{c}{NDCG@5}& \multicolumn{3}{c}{$\rm{mAP@5^\dagger}$} & \multicolumn{2}{c}{$\rm{NDCG@5^\dagger}$}\\
    \cmidrule(l){2-3} \cmidrule(l){4-6} \cmidrule(l){7-8} \cmidrule(l){9-11} \cmidrule(l){12-13} 
    & @0.5 & @0.7 & @0.5 & @0.75 & avg & @0.5 & @0.75 & @0.5 & @0.75 & avg & @0.5 & @0.75 \\
    \midrule
    CLIP-shot~\cite{radford2021clip} & 14.07 & 11.70 & 13.18 & 9.81 & 11.41 & 15.89 & 12.74 & 23.45 & 17.61 & 20.19 & 20.91 & 14.21 \\
    Moment-DETR~\cite{lei2021qvhighlights} & 21.65 & 12.00 & 22.65 & 10.66 & 16.49 & 26.4 & 15.42 & 22.57 & 8.35 & 16.16 & 29.71 & 16.94 \\
    QD-DETR~\cite{moon2023query} & 21.49 & 12.27 & 22.83 & 11.30 & 16.9  & 26.13 & 15.59 & 22.81 & 9.40 & 16.76 & 30.01 & 18.14 \\
    RATV~\cite{yang2023shot} & 3.86 & 1.61 & 3.39 & 1.21 & 2.29  & 3.95 & 1.71 & 6.33 & 1.37 & 4.29 & 14.39 & 5.11 \\
    PREM~\cite{hou2024improving} & 17.35 & 12.95 & 26.51 & 18.27 & 21.88  & 26.25 & 19.49 & 30.74 & 20.71 & 24.87 & 29.35 & 22.60 \\
    \midrule 
    \textbf{TV-MGI (Ours)} & \textbf{30.91} & \textbf{26.14} & \textbf{38.7} & \textbf{30.04} & \textbf{33.39} & \textbf{39.27} & \textbf{32.58} & \textbf{44.52} & \textbf{33.77} & \textbf{37.93} & \textbf{54.82} & \textbf{47.15}\\
    \bottomrule \\
    \end{tabular}}
\caption{Main results on the MSSD test split. R1, mAP@5, NDCG@5 stand for recall@1, mean average precision at five and normalized discounted cumulative gain at five respectively, with higher values indicating better performance. $\rm{mAP@5^\dagger}$ and $\rm{NDCG@5^\dagger}$ denote script level evlautaion. Our approach outperforms competing methods across all evaluation metrics.}
\label{tab:main}
\end{table*}

\begin{table*}
    \centering
    \resizebox{\textwidth}{!}{\begin{tabular}{lcccccccccccccc}
        \toprule
        \multirow{2}{*}{Method} & \multicolumn{2}{c}{R1} & \multicolumn{3}{c}{mAP@5} & \multicolumn{2}{c}{NDCG@5}& \multicolumn{3}{c}{$\rm{mAP@5^\dagger}$} & \multicolumn{2}{c}{$\rm{NDCG@5^\dagger}$}\\
    \cmidrule(l){2-3} \cmidrule(l){4-6} \cmidrule(l){7-8} \cmidrule(l){9-11} \cmidrule(l){12-13} 
    & @0.5 & @0.7 & @0.5 & @0.75 & avg & @0.5 & @0.75 & @0.5 & @0.75 & avg & @0.5 & @0.75 \\
        \midrule
        Ours & \textbf{30.91} & \textbf{26.14} & \textbf{38.7} & \textbf{30.04} & \textbf{33.39} & \textbf{39.27} & \textbf{32.58} & \textbf{44.52} & \textbf{33.77} & \textbf{37.93} & \textbf{54.82} & \textbf{47.15}\\ \addlinespace[0.1cm]
         w/o Shot fusion & 29.08 & 24.28 & 37.23 & 28.24 & 30.76 & 38.81 & 31.77 & 41.52 & 30.36 & 35.32 & 54.6 & 47.09 \\
         w/o Multi. sent. & 23.67 & 19.79 & 31.41 & 23.78 & 27.11 & 31.64 & 25.68 & 36.3 & 28.1 & 21.96 & 46.62 & 38.85\\
         w/o Injection & 28.26 & 23.74 & 36.8 & 28.25 & 31.84 & 37.45 & 30.79 & 38.84 & 30.39 & 33.49& 61.43& 53.32\\
         w/o Ranking loss & 19.57 & 15.71 & 26.43 & 18.8 & 21.97 & 25.86 & 19.86 & 36.19 & 25.96 & 29.9 & 48.08 & 38.91\\
        \bottomrule
    \end{tabular}}
    \caption{Ablation results on the MSSD test split. Each of the last four lines represents a method without that specific strategy. The results show that our strategies are effective, and removing any one of them leads to a performance drop.}
    \label{tab:ablation}
\end{table*}

\subsection{Experimental Settings}
\paragraph{Evaluation metrics.} 
The quality of the generated short video montage is evaluated by assessing the alignment between the predicted video segments and the ground truth segments at both the sentence and script levels. At the sentence level, we employ the following metrics: Recall@1 (R1), mean average precision (mAP), and Normalized Discounted Cumulative Gain (NDCG) at different Intersection-over-Union (IoU) thresholds. The top five predicted segments are selected for each sentence to ensure comprehensive evaluation when computing mAP and NDCG. A prediction is considered positive for R1 and NDCG@5 if its IoU with a ground truth shot exceeds the threshold. Additionally, the average mAP is calculated across a range of IoU thresholds from $0.3$ to $0.95$, with an increment of $0.05$. For the NDCG metric, positive samples are chosen based on an IoU threshold, and the relevance score is assigned as the $rank$ of the matching ground truth shot. For samples without a match, the relevance score is set to $0$.
Similarly, at the script level, all sentences within each script are merged, and the top five predicted shots are retained for each sentence.

\paragraph{Implementation details.}

We use ChineseCLIP~\cite{chinese-clip}, a Chinese version of the CLIP model, to extract features for both video frames and script sentences. Video frames are sampled at a rate of $2$ FPS. We do not input script duration directly into the model. The duration information of the script is implicitly modelled through sentence length and frame sampling (sample in 2 fps), utilizing positional encoding. It is worth noting that we will mask the subtitles in the video frames before inputting them into the model to prevent leakage of subtitle information. We freeze the visual encoder during training while maintaining a trainable text encoder. We configure the fusion encoder with three layers, a hidden size of $512$, and $8$ attention heads. The model is trained for $30$ epochs with a total batch size set to $72$. We use AdamW~\cite{loshchilov2017decoupled} with a cosine learning rate annealing policy and a warm-up ratio of $0.01$. The peak learning rate is set to $1\times{10^{-5}}$ for the text encoder and $1\times{10^{-4}}$ for the rest of the model. And the temperature scaling parameter $\tau$ is set to $0.5$. The experiments are conducted on 8 NVIDIA V100s with 32GB memory per GPU. 

\subsubsection{Baselines}

We use the following works as the baselines to
evaluate our proposed method. \textbf{CLIP-shot} computes the CLIP~\cite{radford2021clip} scores between each sentence and shot, selecting the top four shots with the highest scores as predicted windows. \textbf{Moment-DETR}~\cite{lei2021qvhighlights} and \textbf{QD-DETR}~\cite{moon2023query} considered each sentence in the script as an independent input, resulting in multiple video segments corresponding to that specific sentence. \textbf{RATV} entails a VMG method~\cite{yang2023shot} for multi-sentence queries to retrieve relevant video segments, which were then segmented based on sentence duration. \textbf{PREM} leverages a VCMR method~\cite{hou2024improving} to perform multi-pass, sentence-level queries in order to retrieve relevant video segments.

\subsection{Quantitative Analysis}

Table~\ref{tab:main} shows the main results of different methods on the MSSD test split. Our approach outperforms all baseline methods across all metrics, demonstrating its ability to generate high-quality short video montages with precise alignment between sentences and video segments. Our experiments showed that QD-DETR has a slight advantage over Moment-DETR. However, both methods experience a significant drop in performance at the IoU threshold of 0.7. This can be attributed to these methods' lack of shot-level feature extraction, as they can only process multiple video inputs as a single frame sequence. All baseline methods performed worse than our method on the NDCG metric (PREM 22.60 vs. Ours 47.15). This indicates that existing video moment retrieval methods are not well-suited for handling the complex scenarios in our task. RATV's performance is unsatisfactory, which may be because they only retrieve videos for the overall script, disregarding the temporal synchronization between the script and videos.

\begin{figure*}[!t]
\centering
\includegraphics[trim=0cm 0cm 0 0cm, clip, width=\textwidth]{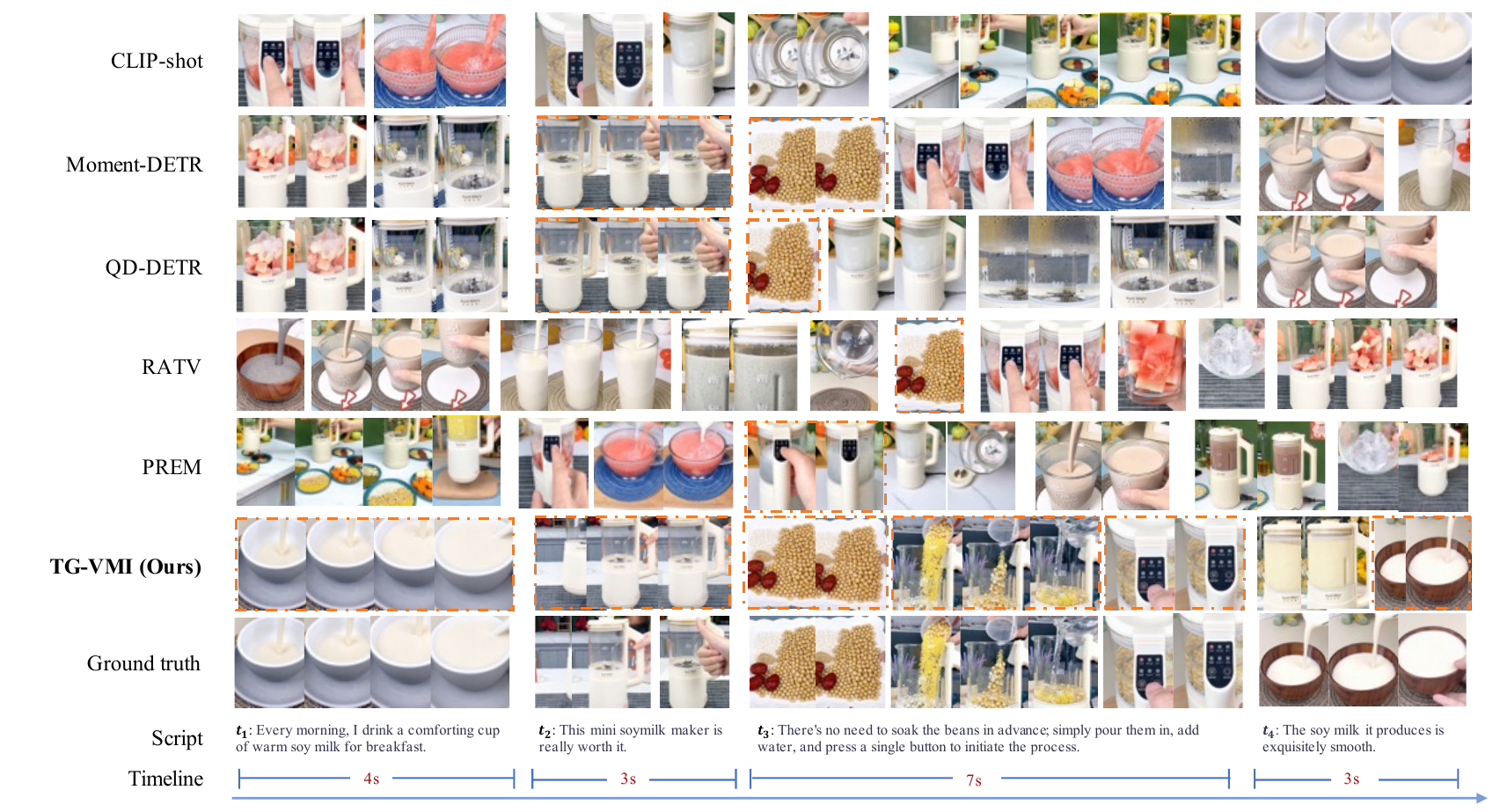}
\caption{An example of short video montages generated by different methods. Four consecutive sentences from a single script and the corresponding video segments generated by each method are presented in separate rows. The orange dashed line represents the recall of correct segments. Our method performs better in terms of both semantic and temporal consistency.}
\label{fig:case}
\end{figure*}

\subsection{Qualitative Analysis}
We provide an example of short video montages from the test set generated by various methods in Figure~\ref{fig:case}. Four consecutive sentences from the script are selected, and the corresponding video segments are displayed. From a semantic matching perspective: 1) The CLIP-based approach lacks precision in matching video shots, especially for the third sentence. 2) The moment retrieval methods struggle with accurate visual ordering when dealing with long sentences, leading to the retrieval of incorrect shots. For instance, Moment DETR fails to depict the pouring of beans and water in the third sentence. 3) RATV primarily focuses on ranking retrieved materials rather than video localization, resulting in poor matching accuracy and temporal alignment performance. In contrast, our method incorporates contextual information, allowing for the retrieval of relevant shots. 4) Our approach achieves the highest recall rates and aligns the order of retrieved segments with the ground truth. Notably, our method successfully matches the scene of soy milk being made and poured into a bowl in the fourth sentence, resulting in a more coherent narrative overall.

\subsection{Ablation Study}
\label{sec:ablation}
Table~\ref{tab:ablation} shows the results of an ablation study conducted on the MSSD test split. 1) \textit{Shot Fusion}. Removing the shot fusion module and relying solely on the final shot-level feature aggregation significantly decreased retrieval performance. Explicitly modelling shot-level representations enhances the models' discriminative capabilities and improves shot differentiation proficiency. 2) \textit{Multi. sent.} Disregarding the contextual relationships between script sentences and treating them as independent inputs significantly decreased performance, emphasizing the importance of modelling the relationships between multiple script sentences.
 3) \textit{Injection}. Removing the injection strategy for synthetic data augmentation also decreased performance, indicating the strategy's role in bridging the gap between training data and real-world scenarios. 4) \textit{Ranking loss}. 
The ranking loss module significantly improved model performance, elevating the NDCG metric from 38.91 to 47.15. The results show that this module effectively distinguishes between positive and negative samples, enhancing the model's discriminative capabilities and enabling efficient shot ranking within each sentence. Our method outperforms other approaches even without the ranking loss, owing to the effectiveness of the multi-grained fusion encoder in facilitating efficient cross-modal interactions and capturing the underlying positional information of sentences and shots.
In summary, each component significantly contributes to the method's overall performance.

\subsection{User Study}
\begin{figure}[!t]
\centering
\includegraphics[trim=0 2 0 0,clip,width=0.4\textwidth]{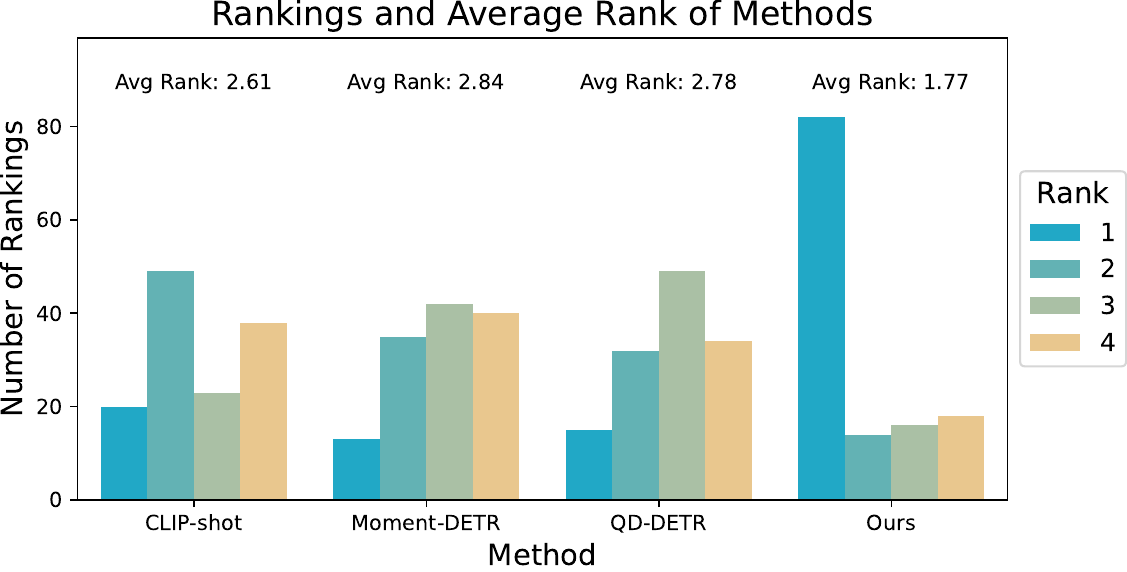}
\caption{User study results. For a given script, users ranked videos generated by four different methods, where rank 1 is best, and rank 4 is worst. We report the frequency of each rank received by each method and the average rank for each.}
 %
\label{fig:user_study}
\end{figure}

We conduct a user study to evaluate the performance of four different methods of video generation. We randomly selected 50 scripts from the MSSD test split for this study and generated 200 videos using the four methods. We then invite volunteers to rank these videos based on the semantic and temporal coherence between the generated videos and the corresponding scripts. 
Our volunteer group consists of 13 individuals, and each volunteer is asked to rank ten video sets. All of our volunteers have extensive experience using short video platforms and some experience in video editing. Consequently, each set of data has rankings from two different individuals. Furthermore, the visual results produced by RATV and PREM exhibit a chaotic nature, as depicted in Figure~\ref{fig:case}. To avoid misleading volunteers’ judgments, we remove their results from the user study. The results are presented in Figure~\ref{fig:user_study}. Our method significantly outperforms other methods in terms of perceptual experience, providing strong evidence for the effectiveness of our approach.


\section{Conclusion}

In this work, we introduce a novel task called \task. Given a multi-sentence script and multiple source videos, this task aims to automatically localize video segments for each sentence to create a short video montage. To tackle this task, we propose a multi-grained multimodal transformer framework capable of fusing multimodal features at various levels to facilitate improved comprehension and more effective montage generation. Furthermore, we collect a large-scale dataset for this task and present experiments demonstrating our approach's efficacy. 
\bibliography{aaai25}

\end{document}